\documentclass[journal,twoside,web]{ieeecolor}
\usepackage{generic}
\usepackage{cite}
\usepackage{amsmath,amssymb,amsfonts}
\usepackage{algorithmic}
\usepackage{graphicx}
\usepackage{algorithm,algorithmic}
\usepackage{hyperref}
\usepackage{textcomp}
\usepackage{booktabs}
\usepackage{multirow}
\usepackage{subcaption}
\usepackage[margin=1in]{geometry}

\def\BibTeX{{\rm B\kern-.05em{\sc i\kern-.025em b}\kern-.08em
    T\kern-.1667em\lower.7ex\hbox{E}\kern-.125emX}}

\begin{document}
\title{Rethinking Suicidal Ideation Detection: A Trustworthy Annotation Framework and Cross-Lingual Model Evaluation}
\author{Amina Dzafic, Merve Kavut, and Ulya Bayram\\
Department of Electrical and Electronics Engineering,\\
Canakkale Onsekiz Mart University,\\
Canakkale, 17110, Türkiye.\\
\texttt{ulya.bayram@comu.edu.tr}
}

\maketitle

\begin{abstract}
Suicidal ideation detection is critical for real-time suicide prevention, yet its progress faces two under-explored challenges: limited language coverage and unreliable annotation practices. Most available datasets are in English, but even among these, high-quality, human-annotated data remains scarce. As a result, many studies rely on available pre-labeled datasets without examining their annotation process or label reliability. The lack of datasets in other languages further limits the global realization of suicide prevention via artificial intelligence (AI). In this study, we address one of these gaps by constructing a novel Turkish suicidal ideation corpus derived from social media posts and introducing a resource-efficient annotation framework involving three human annotators and two large language models (LLMs). We then address the remaining gaps by performing a bidirectional evaluation of label reliability and model consistency across this dataset and three popular English suicidal ideation detection datasets, using transfer learning through eight pre-trained sentiment and emotion classifiers. These transformers help assess annotation consistency and benchmark model performance against manually labeled data. Our findings underscore the need for more rigorous, language-inclusive approaches to annotation and evaluation in mental health natural language processing (NLP) while demonstrating the questionable performance of popular models with zero-shot transfer learning. We advocate for transparency in model training and dataset construction in mental health NLP, prioritizing data and model reliability.
\end{abstract}


\noindent\textbf{Note:} This manuscript has been submitted to the IEEE Journal of Biomedical and Health Informatics.

\section{Introduction}

Suicide is a serious global public health issue, which is the third leading cause of death among 15 to 29-year-olds worldwide \cite{who_suicide_2025}. About 70\% of individuals with suicidal ideation do not receive professional help due to stigma and lack of access to care, among other factors \cite{kessler2005prevalence, kazdin2011evidence}. A solution considered by many has been to adopt recent developments in AI for suicide prevention \cite{ji2020suicidal}. Numerous studies have applied machine learning and deep learning methods to data collected in clinical settings \cite{riera2024clinical, pigoni2024machine}. While these clinical data collections are the gold standard, obtaining them is costly and subject to strict regulations \cite{ji2020suicidal}. Researchers who lack these resources and those aiming to capture more authentic expressions of suicidal ideation that may remain hidden in clinical settings have turned to social media platforms as data sources, such as Twitter (now X) \cite{coppersmith2018natural, shukla2025enhancing}, followed by Reddit with subreddits like r/SuicideWatch \cite{gorai2024bert, qorich2024advanced}. Studies employed NLP techniques ranging from traditional word occurrence statistics to deep learning and transformer-based models. However, the reliance on English data limits the global relevance of these studies, particularly in non-English-speaking regions where suicide prevention is equally vital.

Türkiye is one of the countries where suicide is a critical public health concern. A study computed a 304.7\% increase in suicide rates over the 38 years until 2017 \cite{yakar2017suicide}. Another demographic investigation reported that the highest suicide rates were among adults over the age of 75 \cite{erenler2023suicide}. Suicide has been at concerning rates among Turkish medical students and physicians as well \cite{yildiz2023suicide}. Based on ongoing socioeconomic changes in the country, these rates are expected to rise across different demographics \cite{yucel2023suicidal}. All of these statistics underscore the urgency of suicide prevention in Türkiye. However, AI research for Turkish suicidal ideation is nearly non-existent. Because the Turkish language lacks annotated suicidal ideation detection corpora, and translating existing English datasets is unfeasible due to the non-transferability of cultural and linguistic nuances. Consequently, there is a critical need to collect and annotate Turkish social media data for suicidal ideation, a resource-intensive process. One of our objectives is to construct a novel Turkish dataset on suicidal ideation, ensuring reliable annotations with minimal resources: two researchers and an expert annotator. To augment our resources, we propose leveraging LLMs in the annotation process without compromising reliability.

In contrast with Turkish NLP, there is an abundance of studies in English suicidal ideation detection and several popular social media datasets with labels \cite{gaur2019knowledge, desu2022suicide, ji2021suicidal}. To develop a Turkish suicidal ideation detection framework informed by the studies with the same objective conducted in English, we raise two critical questions: Can we trust the dataset labels? And, can we trust the models trained or fine-tuned on these datasets? We address these questions through empirical analysis and context-sensitive evaluation in Turkish and English. We evaluate our annotated Turkish corpus alongside three widely used English Reddit datasets labeled for suicidal ideation. With transfer learning, we obtain sentiment and emotion classifications from the labeled posts of the datasets. We not only use these models to evaluate the quality and consistency of annotations but also to examine model performance on datasets with human-validated (gold-standard) labels, thereby allowing bidirectional insight into both annotation and model reliability.

In the remainder of this paper, we review related work, describe our dataset and annotation process, and present the selected English corpora and the transformers. We then continue with our evaluations and discuss the implications for both data reliability and ethical implications.

\section{Related Work}

Recent studies, in harmony with the rest of the NLP literature, use deep learning and transformers in identifying suicidal ideation from social media. One study used a public Twitter sentiment dataset, and a Reddit dataset consisting of posts from r/SuicideWatch and r/Depression subreddits automatically labeled as suicidal \cite{gorai2024bert}. They proposed an ensemble model consisting of Bidirectional Encoder Representations from Transformers (BERT) and multiple Convolutional Neural Networks (CNNs), achieving a 97.1\% accuracy on Reddit and a 99.4\% on Twitter. Another study used two datasets: the Suicide and Depression Detection (SDD) Reddit dataset and the Twitter Suicidal Intention Dataset, both with binary labels for suicidal ideation \cite{shukla2025enhancing}. After experimenting with machine learning and deep learning methods, they achieved F1 scores of 97\% on Reddit and 98\% on Twitter. An alternative study using the same SDD dataset reported a 94.9\% F1 score \cite{qorich2024advanced}. Likewise, a study using the same SDD dataset and another Reddit collection containing SuicideWatch and Mental Health (SWMH) subreddit posts achieved F1 scores of 97\% on SDD and 68\% on SWMH \cite{ezerceli2024mental}. All these high scores in detecting suicidal ideation have influenced many researchers to pursue improving these scores to perfection. Yet, the datasets that returned these scores remain under-explored for label reliability.

After an extensive literature search, we found two studies focusing on Turkish suicidal ideation detection. One study surveyed young adults aged 18–30 using questionnaires \cite{turk2023predicting}, and predicted those who had suicidal ideation within the past year with a 75\% accuracy. The other study collected 25,000 Turkish tweets containing mental health keywords and automatically labeled them \cite{alshammari2024mental}. After applying sentiment lexicons and manually reviewing a subset of the labels, they fine-tuned a Turkish BERT model for three-class sentiment analysis, achieving an accuracy of 82.6\%. While these efforts are positive steps toward suicide prevention in Turkey, further advancements are required. We aim to contribute to this progression in Türkiye and to develop more reliable, data-driven, global mental health interventions.

\section{Data}

\subsection{The Novel Turkish Dataset}

We selected Ek\c{s}i S\"{o}zl\"{u}k (\url{https://eksisozluk.com/}) as our social media domain because it is older than Twitter, solely text-based, and has been one of the prominent platforms in Turkey \cite{sari2025vaccine}. Unlike Reddit, authors do not respond to each other. We selected the six discussion topics listed in Table~\ref{tab:Turkish_dataset_info} and collected their posts from May 4, 1999, to March 6, 2025, through the BeautifulSoup web scraper in Python \cite{richardson2007beautiful}. In total, we obtained 7,874 Turkish social media posts. Table~\ref{tab:Turkish_dataset_info} shows that the suicide topic with the highest number of authors and posts is ``suicide,'' followed by the potential suicide notes and understanding those who committed suicide. These topics also have high average post lengths.

\begin{table}[t!]
\centering
\caption{The selected topics, their numbers of posts and authors, and average post lengths with standard deviations computed as the number of words.}
\begin{tabular}{p{1.75in}p{.1in}p{.27in}p{.38in}}
\toprule
\textbf{Topic/Translation} & \textbf{Posts} &  \textbf{Authors} & \textbf{Lengths} \\
\midrule
hiç yaşamamış olmayı ister miydin (would you prefer never existing) & 210 & 210 & 30 (41) \\
intihar (suicide) & 4525 & 3,498 & 88 (168) \\
intihar eden bir insanı anlamak (understanding one who committed suicide) & 1193 & 1,097 & 73 (134) \\
intiharı düşünmek (thinking of suicide) & 645 & 608 & 54 (79) \\
kendini asmak (hanging yourself) & 154 & 153 & 56 (80) \\
sözlük yazarlarının olası intihar notları (potential suicide notes) & 1147 & 1,090 & 27 (101) \\
\bottomrule
\end{tabular}
\label{tab:Turkish_dataset_info}
\vspace{-15pt}
\end{table}

\begin{figure*}[t!]
\begin{subfigure}[b]{.78\textwidth}
    \includegraphics[width=\textwidth]{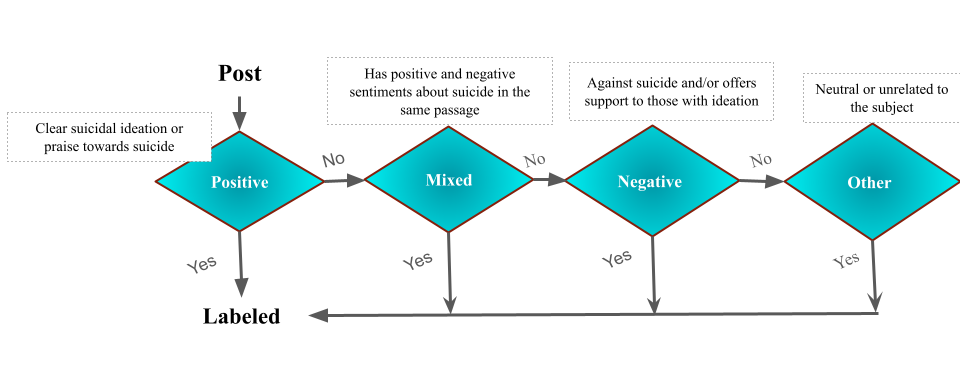}
    \caption{Manual annotation framework}
    \label{fig:framework}
\end{subfigure}
\begin{subfigure}[b]{.21\textwidth}
    \includegraphics[width=\textwidth]{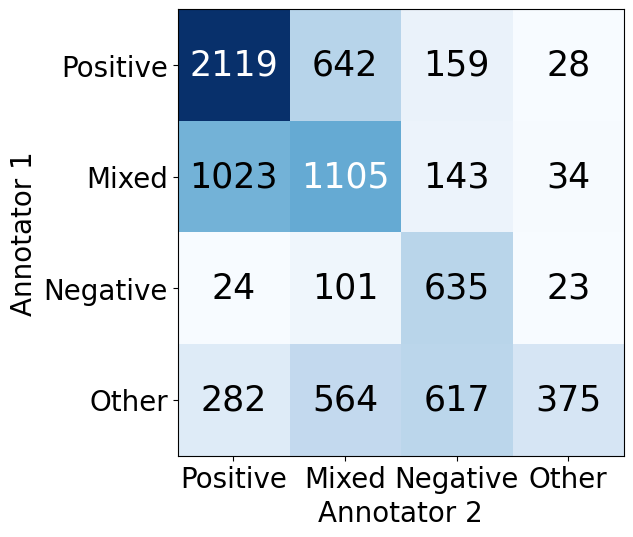}
    \caption{Annotator agreement}
    \label{fig:two_annotator_conf}
\end{subfigure}
\caption{Our labeling framework and resulting annotator confusion matrix.}
\end{figure*}

The available resources allowed two researchers and one domain expert to participate in the annotation process. After collecting the data, we collaboratively identified four main labels and developed the set of annotation guidelines in Fig.~\ref{fig:framework}. For each post, two researchers determined whether it had suicidal ideation. If it had unequivocal suicidal thoughts without discouraging others, it is labeled Positive. If suicidal ideation is present with discouragement, its label is Mixed. The remaining two labels lack suicidal ideation: The Negative label opposes suicide and promotes help, and the Other remains neutral or talks about unrelated subjects. The confusion matrix of label assignments by the two annotators is in Fig.~\ref{fig:two_annotator_conf}, showing 54\% agreement, which is not acceptable to finalize the annotation.

\begin{figure}[t!]
    \centering
    \includegraphics[width=1\linewidth]{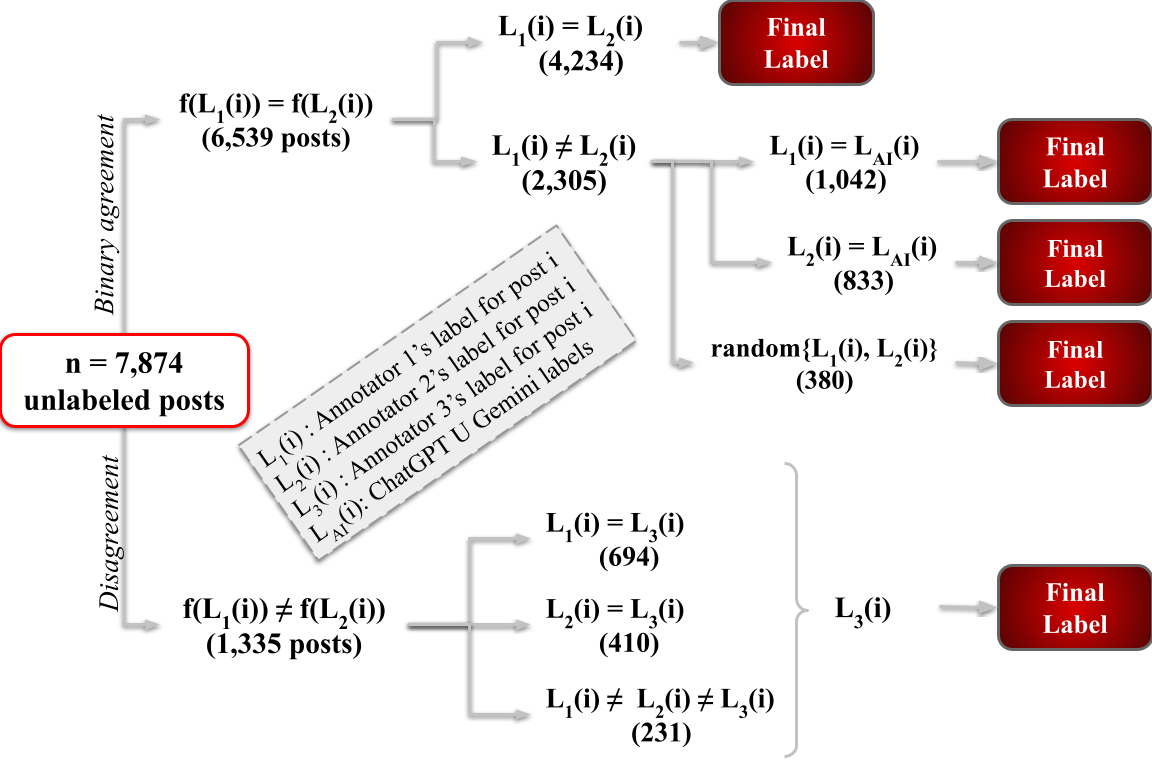}
    \caption{The decision-making process for assigning the final labels to the Turkish corpus.\vspace{-15pt}}
    \label{fig:decision}
\end{figure}

To handle the two annotator disagreements, we developed the decision-making strategy in Fig.~\ref{fig:decision}. First, we group the four labels into a binary problem:
\begin{equation}
    f(x) = 
    \begin{cases} 
      1 & \text{if } x \in \{Positive, Mixed\} \\
      0 & \text{if } x \in \{Negative, Other\}
    \end{cases}
    \label{eq:binarize}
\end{equation}
\noindent based on the presence of suicidal ideation in a post. Next, we separate the disagreement posts into two. Two annotators completely disagree about the binary label of 1,335 posts, while agreeing the binary label of 2,305 posts. For the disagreement cases with a binary agreement, we used ChatGPT-4.o and Gemini 2.5 as free LLM labeling tools. Recent studies have evaluated their capabilities in text annotation. One study reported 84\% accuracy for ChatGPT's annotation performance across ten English datasets \cite{aldeen2023chatgpt}. Another study reported a 92\% agreement between ChatGPT and human annotations in Turkish \cite{ezin2024using}, yet a study claimed human annotations to be better \cite{nasution2024chatgpt}. So, instead of relying on the automated labels generated by these LLMs, we constructed the rule in Fig.~\ref{fig:decision}: Exact agreement between an LLM and a human annotation become the final label. Only 380 posts remain without any agreement between the annotators and LLMs. We turn these cases into advantages by assigning them a random label out of human annotations. Therefore, we introduce a small noise into the dataset; and we do so without affecting its the binary label.

The 1,335 posts with complete disagreement are the cases where two annotators disagree about the presence of suicidal ideation in them. LLMs cannot be used in these sensitive labeling problems, so we used the expert opinion as the third human annotator. As Fig.~\ref{fig:decision} shows, the majority of these posts have a label agreement between the expert and one of the two annotators, with only 231 posts labeled exclusively by the expert.

\begin{table}[h!]
\centering
\caption{Statistics of the labeled Turkish dataset as the number of authors, posts, and average (standard deviation) post lengths grouped by final labels.}
\begin{tabular}{lccc}
\hline
\textbf{Labels} & \textbf{Authors} & \textbf{Posts} & \textbf{Lengths} \\
\hline
Positive & 2,529 & 2,985 & 55 (121) \\
Mixed    & 2,118 & 2,429 & 85 (156) \\
Negative & 1,213 & 1,301 & 80 (129) \\
Other    & 1,073 & 1,154 & 52 (143) \\
\hline
\end{tabular}
\label{tab:tr_dataset_stats}
\end{table}

Table~\ref{tab:tr_dataset_stats} shows the complete collection has 6,078 authors. We observe that most authors have one post per topic, suggesting a diverse contributor collection. Only 311 authors, have more than one post in the complete set. Of these individuals, 301 have the same final label on their multiple posts. This consistency supports the validity of our annotation framework, consistently capturing the persistent mindset of the same author across topics. The remaining ten authors with multiple posts have different labels, which is an acceptably small noise. Table~\ref{tab:tr_dataset_stats} further shows that most of the authors contain suicidal ideation at varying levels, while neutral posts are in the minority. Another notable observation concerns average post lengths. Authors with Mixed opinions have the longest posts, reflecting their internal conflict, followed by those expressing Negative views towards the subject, persuading others against suicide.

\subsection{English Datasets}

We selected three prominent, non-overlapping Reddit datasets, and excluded Twitter for its word limit.

\textbf{C-SSRS} was constructed by scraping 8 million posts from 270,000 authors and 15 mental health subreddits (e.g., r/selfharm, r/bipolar, r/Anxiety, r/SuicideWatch, r/schizophrenia, r/depression, among others) \cite{gaur2019knowledge}. After eliminating authors without r/SuicideWatch subreddit posts and randomly selecting 500 users from the remaining, they obtain a collection spanning years from 2005 to 2016, and gets them labeled by four practicing psychiatrists with five labels in Table~\ref{tab:cssrs}. Suicidal Ideation posts have thoughts of suicide and preoccupations with risk factors, Suicidal Behavior posts have a higher suicide risk with a confession of active or historical self-harm or planning, Actual Attempt posts have intentional actions that could result in death. As non-suicidal labels, they define Suicide Indicator label for supportive posts that use at-risk words from a clinical lexicon, and Supportive label for the remaining posts. Throughout the process, the psychiatrists used the Columbia Suicide Severity Rating Scale (C-SSRS) guidelines with a labeler agreement of 73\%.
This collection is used by various studies, achieving F1 scores in correctly classifying the five labels, ranging from 65\% to 79\% \cite{gaur2019knowledge,naseem2022hybrid,soun2024rise}. Table~\ref{tab:cssrs} shows that the majority of the authors have suicidal ideation, while the supportive labels have fewer authors. Likewise, most posts have the Ideation label, followed by Supportive posts. Yet, average post lengths are similar, with the longest posts from the Attempt and Indicator categories.

\textbf{SDD} dataset originally consisted of posts from r/SuicideWatch, r/depression, and r/teenagers subreddits \cite{desu2022suicide}. The version we use remove the depression posts, with resulting SDD spanning December 15, 2008, to January 1, 2021 \cite{nikhileswar2021suicide}. Neither version involves human annotations. According to Table~\ref{tab:sdd_dataset}, suicidal posts are lengthier than non-suicidal ones. Before publicizing the dataset, they exclude information identifying the authors. Studies use either version of this dataset, achieving 94.95\% to 97\% F1 scores \cite{shukla2025enhancing,qorich2024advanced,lin2024data,ezerceli2024mental}.

\textbf{SWMH} dataset also consists of various mental health subreddits discussing depression, anxiety, and bipolar disorders, among others, and is available from Hugging Face (datasets/AIMH/SWMH) \cite{ji2021suicidal}. It contains 54,412 posts. Table~\ref{tab:swmh_dataset} shows that its labels are subreddit titles and the author information is absent. The highest number of posts belongs to the depression subreddit, followed by SuicideWatch. Offmychest has the longest average post lengths, followed by SuicideWatch. Studies using this collection report 64.78\% and 68\%  F1 scores \cite{ji2021suicidal,ezerceli2024mental}.

\begin{table}[t!]
\centering
\caption{Summary of the selected Reddit-based suicidal ideation datasets.}
\subcaptionbox{Label-level summary statistics for the C-SSRS dataset.\label{tab:cssrs}}[1.0\linewidth]{%
\begin{tabular}{llll}
\toprule
\textbf{Labels}        & \textbf{Authors} & \textbf{Posts} & \textbf{Lengths} \\
\midrule
Ideation     & 171     & 2,746 & 74 (92) \\
Supportive   & 108     & 2,051 & 68 (92) \\
Indicator    & 99      & 1,789 & 79 (106) \\
Behavior     & 77      & 1,789 & 66 (95) \\
Attempt      & 45      & 752   & 89 (114) \\
\bottomrule
\end{tabular}
}
\par\vspace{0.6em}
\subcaptionbox{Label-level summary statistics for the SDD dataset.\label{tab:sdd_dataset}}[1.0\linewidth]{%
\begin{tabular}{lll}
\toprule
\textbf{Labels}         & \textbf{Posts}   & \textbf{Lengths} \\
\midrule
suicide    & 116,037  & 202 (255) \\
non-suicide  & 116,037  & 61 (139) \\
\bottomrule
\end{tabular}
}
\par\vspace{0.6em}
\subcaptionbox{Label-level summary statistics for the SWMH dataset.\label{tab:swmh_dataset}}[1.0\linewidth]{%
\begin{tabular}{lll}
\toprule
\textbf{Labels}         & \textbf{Posts}   & \textbf{Lengths} \\
\midrule
depression    & 18,746  & 159 (221) \\
SuicideWatch  & 10,182  & 182 (243) \\
Anxiety       & 9,555   & 160 (186) \\
offmychest    & 8,284   & 250 (320) \\
bipolar       & 7,645   & 164 (198) \\
\bottomrule
\end{tabular}
}
\vspace{-15pt}
\end{table}

\section{Models}

\begin{table}[t!]
\centering
\caption{The selected transformers and their details.}
\subcaptionbox{Turkish transformers.\label{tab:tr_berts}}[1.0\linewidth]{%
\begin{tabular}{p{.44in}p{1.2in}p{.44in}p{.4in}}
\toprule
Model & Labels & Datasets & Downloads\\
\midrule
MULTI-1  & Two negative, two positive-levels, neutral & Synthetic & 113k \\
MULTI-2 & Two negative, two positive-levels, neutral & Sentences & 694 \\
DistilBERT  & Sadness, joy, love, anger, fear, surprise & Translated dataset & 157  \\
BERTurk-TREMO &  Sadness, happiness, disgust, anger, fear, surprise & TREMO & 438 \\
\bottomrule
\end{tabular}}
\par\vspace{.6em}
\subcaptionbox{English transformers.\label{tab:en_berts}}[1.0\linewidth]{%
\begin{tabular}{p{.55in}p{.55in}p{1in}p{.4in}}
\toprule
Model & Labels & Dataset & Downloads \\
\midrule
SENTINET & suicide/not & C-SSRS, SDD & 326 \\
RoBERTa & suicide/not & SDD & 30.5k \\
EmoRoBERTa & 28 emotions & GoEmotions & 11.3k \\
BERT-Emo & 43 emotions & GoEmotions, Twitter & 612 \\
\bottomrule
\end{tabular}}
\vspace{-15pt}
\end{table}

From the pool of fine-tuned transformers, we selected four Turkish and four English models from Hugging Face in Table~\ref{tab:tr_berts} and Table~\ref{tab:en_berts}. Our selection was guided by both model diversity and popularity (e.g., download counts), ensuring relevance for NLP researchers in model and dataset selection.

The first two models in Table~\ref{tab:tr_berts} are multilingual, fine-tuned to detect positive and negative sentiments (i.e., very negative, negative, very positive, positive) and neutral in various languages, including Turkish. We called the first one MULTI-1 (tabularisai/multilingual-sentiment-analysis), which contains a DistilBERT fine-tuned on synthetic data from multiple sources (e.g., JW300, Europarl, TED Talks) in different languages. The second one is MULTI-2 (agentlans/multilingual-e5-small-aligned-sentiment), based on BERT and fine-tuned on the Multilingual Parallel Sentences dataset in English and various other languages. The third model is another DistilBERT pre-trained on BERTurk data and fine-tuned on a Twitter dataset translated to Turkish via Google API (zafercavdar/distilbert-base-turkish-cased-emotion). For comparison, we include BERTurk-TREMO (coltekin/berturk-tremo), built on BERTurk, a pre-trained Turkish transformer known for its success in Turkish NLP studies \cite{sari2025vaccine}. BERTurk-TREMO is fine-tuned on a Turkish emotion detection dataset. The download counts of the non-multilingual Turkish models demonstrate the scarcity of Turkish NLP studies.

We selected the first two models in Table~\ref{tab:en_berts} for their ability to detect suicidal ideation. The SENTINET (sentinet/suicidality) model is based on ELECTRA and was fine-tuned on various social media datasets, including the full C-SSRS dataset, 7\% of the SDD datasets (with 8,532 matching posts), and several Twitter datasets. The second model, RoBERTa (vibhorag101/roberta-base-suicide-prediction-phr), was fine-tuned on a Reddit dataset with only 216 post overlaps with SDD. The remaining two datasets (C-SSRS and SWMH) have no matches. The remaining two models are fine-tuned for emotion detection in English using the same GoEmotions dataset containing Reddit posts: EmoRoBERTa (arpanghoshal/EmoRoBERTa) and BERT-Emo (borisn70/bert-43-multilabel-emotion-detection). The last model was also fine-tuned on Twitter data. We found no considerable post overlap between the GoEmotions dataset and any of the three benchmark datasets (SDD, C-SSRS, and SWMH), except for a few short and generic posts.

\section{Evaluation Results}

\begin{figure}[t!]
    \centering
    \begin{subfigure}[b]{0.49\linewidth}
    \includegraphics[width=\textwidth]{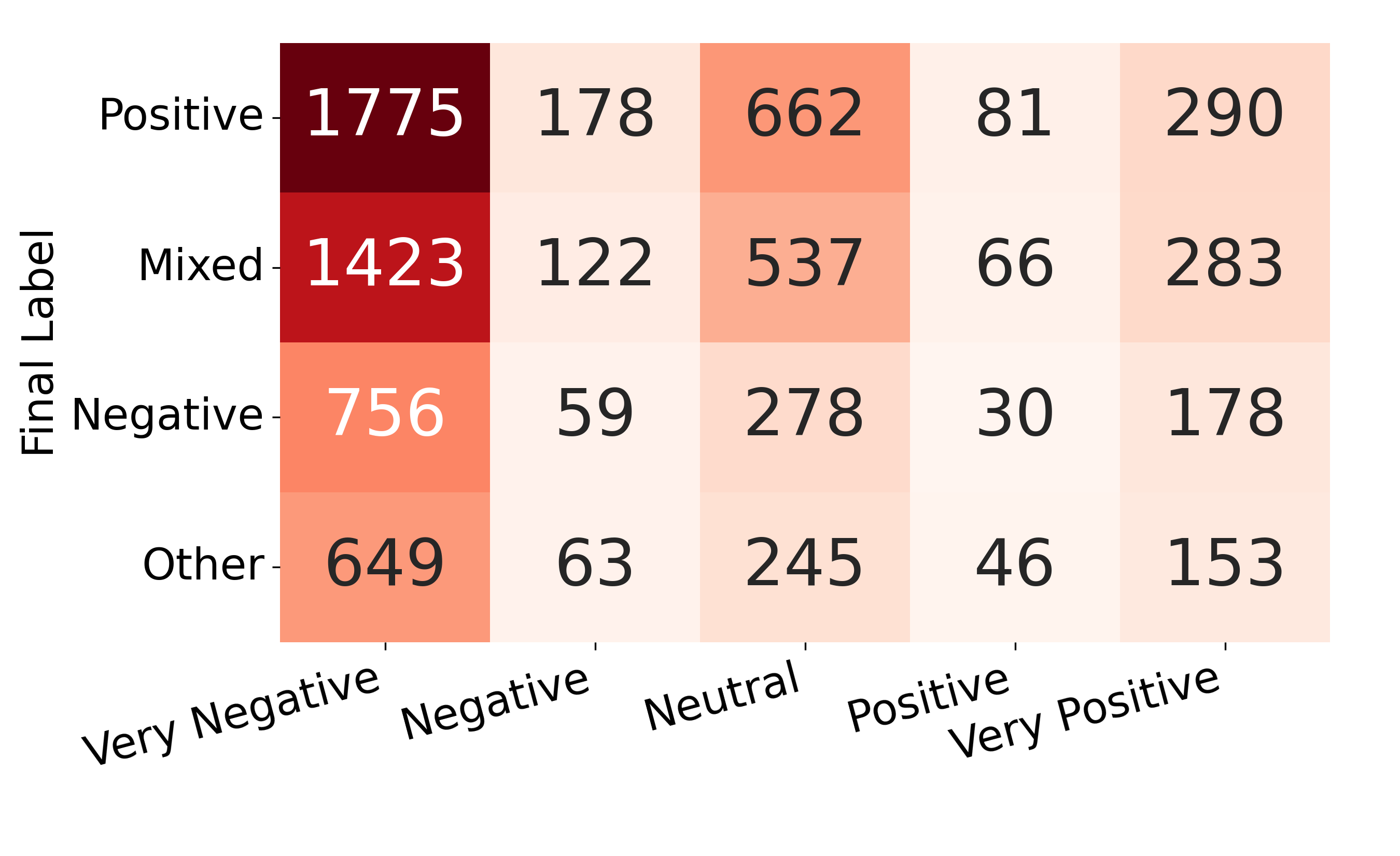} 
    \caption{MULTI-1}
    \label{fig:tr_multi1}
    \end{subfigure}
    \begin{subfigure}[b]{0.49\linewidth}
    \includegraphics[width=\textwidth]{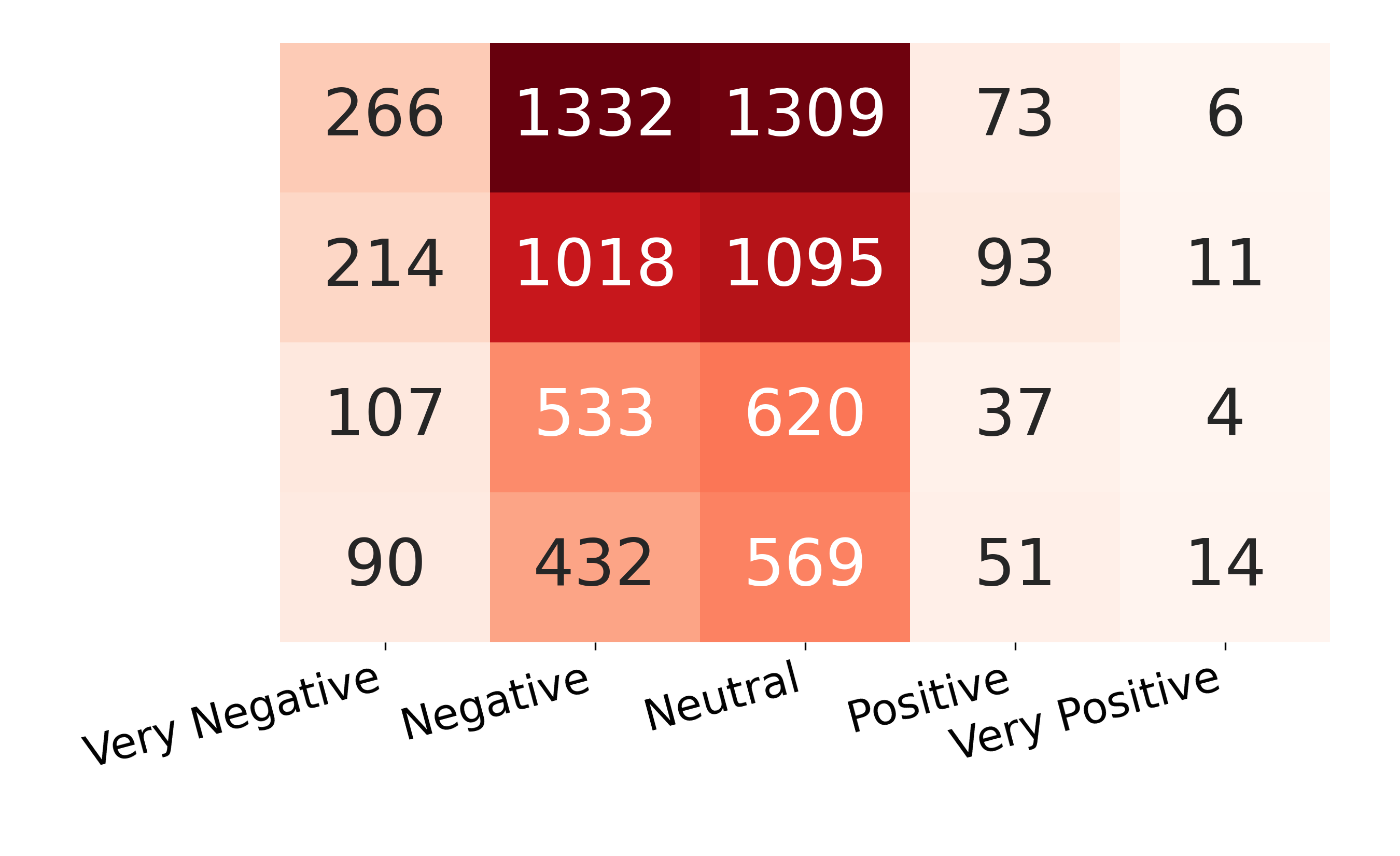} 
    \caption{MULTI-2}
    \label{fig:tr_multi2}
    \end{subfigure}
    \begin{subfigure}[b]{0.48\linewidth}
    \includegraphics[width=\textwidth]{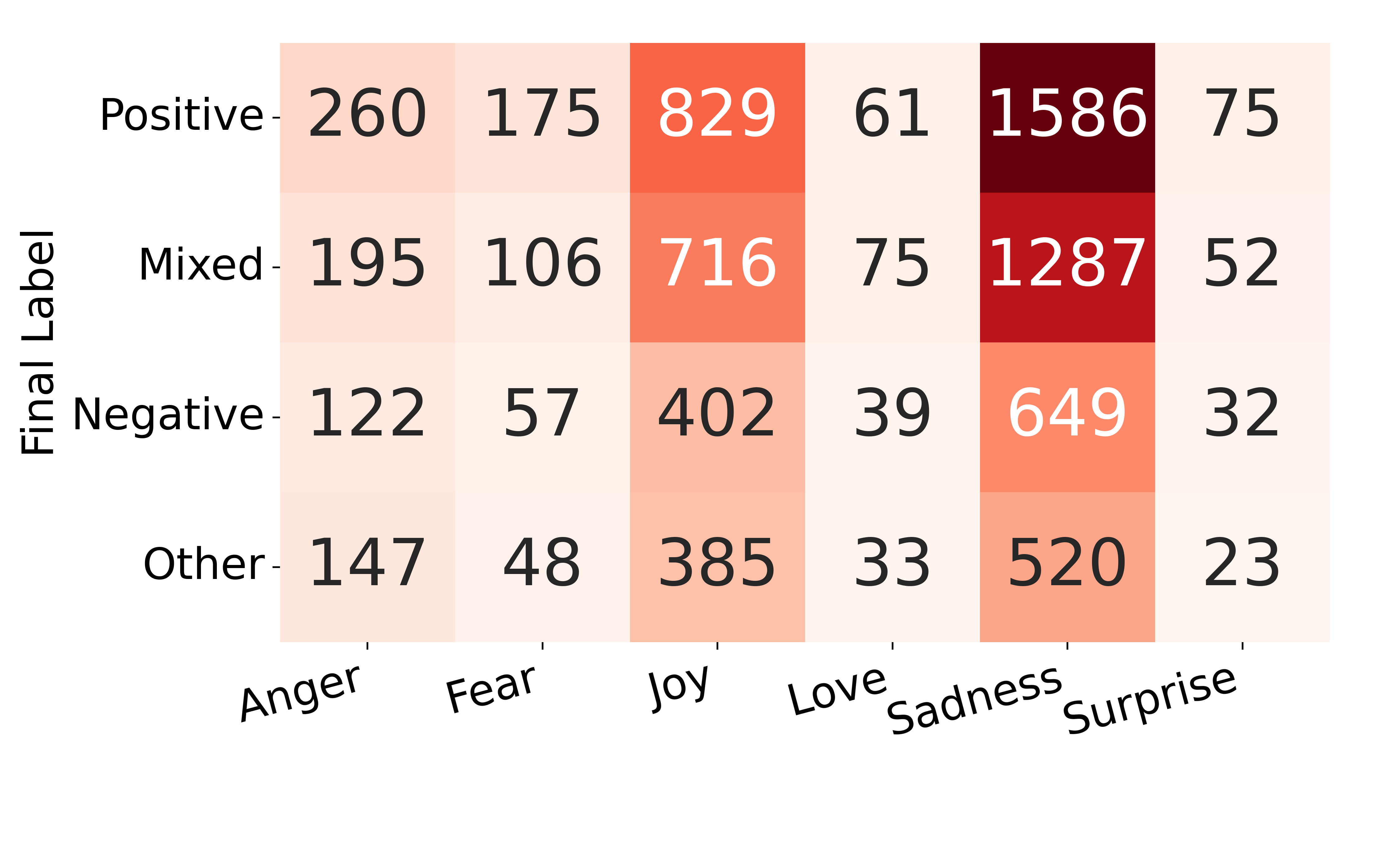} 
    \caption{DistilBERT}
    \label{fig:tr_distil}
    \end{subfigure}
    \begin{subfigure}[b]{0.48\linewidth}
    \includegraphics[width=\textwidth]{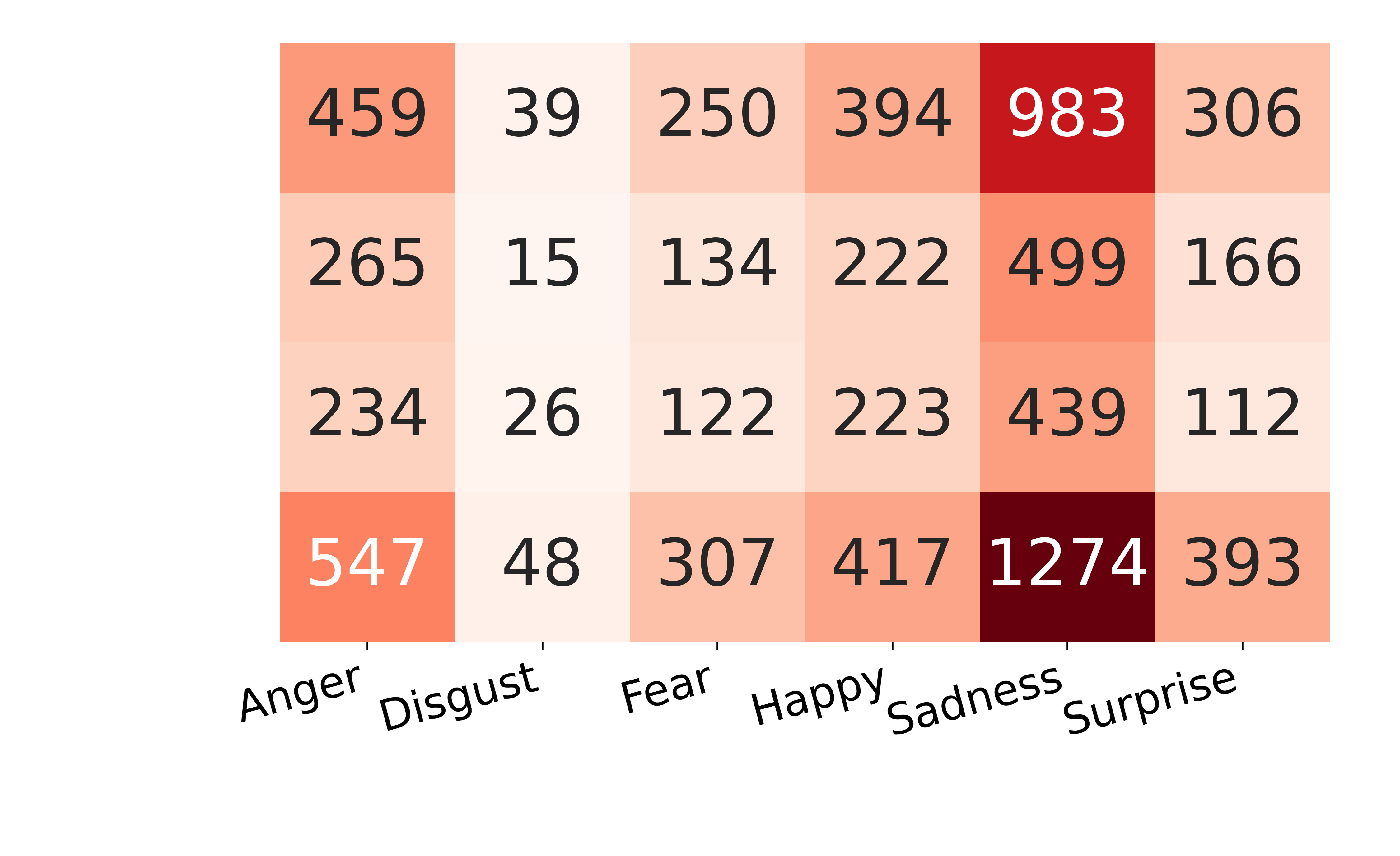} 
    \caption{BERTurk-TREMO}
    \label{fig:tr_emo}
    \end{subfigure}
    \caption{Turkish annotation evaluations.\vspace{-15pt}}
    \label{fig:tr_bert_eval}
\end{figure}

\subsection{Evaluating the Turkish Dataset}

The first evaluation focuses on the novel Turkish suicidal ideation dataset. After obtaining the sentiments and emotions from social media posts using four transformers in Table~\ref{tab:tr_berts}, we selected confusion matrices as the initial metric for evaluations. The first matrix in Fig.\ref{fig:tr_bert_eval} shows that MULTI-1 classified the majority of the Turkish posts as very negative while classifying posts with suicidal ideation as neutral. The confusion matrix of MULTI-2 detected negative and neutral sentiments regardless of the annotation labels. Thus, two sentiment models failed to capture the contextual difference between suicidal and non-suicidal posts. To enumerate the differences between these two multilingual sentiment models, we used the Matthews' Correlation Coefficient (MCC) metric in a multiclass setting, comparing how posts received the same sentiments and obtained an MCC of 0.098. An MCC value close to +1 means agreement, close to 0 means lack of correlation, and -1 means disagreement between two models \cite{stoica2024pearson}.

Next, we evaluate the two Turkish emotion detection models and our dataset. Fig.\ref{fig:tr_bert_eval} DistilBERT detects sadness in the majority of the posts, followed by joy, regardless of their annotations. In contrast, BERTurk-EMO detects a mix of emotions, including anger, love, and surprise, while sadness is prevalent in neutral and positive labels. Although due to the presence of different emotions between the two transformers, we cannot directly compare them, we could compare their emotion detections on the overlapping classes, which returned a 0.241 MCC score. The four models' comparisons returning two MCC scores close to zero demonstrate that we need to beware of the Turkish transformers before relying on their sentiment or emotion classifications.

\subsection{Evaluating the English Datasets}

In evaluating the three English datasets, where only C-SSRS had human annotations, we used two suicidal ideation detection transformers and two emotion detection models. Since the two suicidal transformers, SENTINET and RoBERTa, perform binary classification, we further utilized the macro F1 score and Area Under the ROC Curve (AUC) metrics in evaluations, in addition to confusion matrices and MCC values we employed in the Turkish data evaluation. Since the two emotion detection models span a wide range of emotions, we selected an emotion subset present in both models that could be relevant to suicidal ideation.

\begin{table}[t!]
    \centering
    \caption{Suicidal ideation detection with transformers on selected three Reddit datasets.}
    \begin{tabular}{lcccccc}
    \toprule
        \multirow{2}{*}{Model} & \multicolumn{2}{c}{C-SSRS} & \multicolumn{2}{c}{SDD} & \multicolumn{2}{c}{SWMH} \\
                 & F1  & AUC & F1 & AUC & F1 & AUC \\
        \midrule
        SENTINET & 54.7 & 57.3 & 96.6 & 99.5 & 45.3 & 82.4 \\
        RoBERTa  & 16.3 & 55.8 & 96.9 & 99.5 & 59.5 & 81.0 \\
        \bottomrule
    \end{tabular}
    \label{tab:suicide_res}
    \vspace{-12pt}
\end{table}

\begin{figure}[t!]
    \centering
    \begin{subfigure}[b]{0.36\linewidth}
    \includegraphics[height=.85in, width=\linewidth]{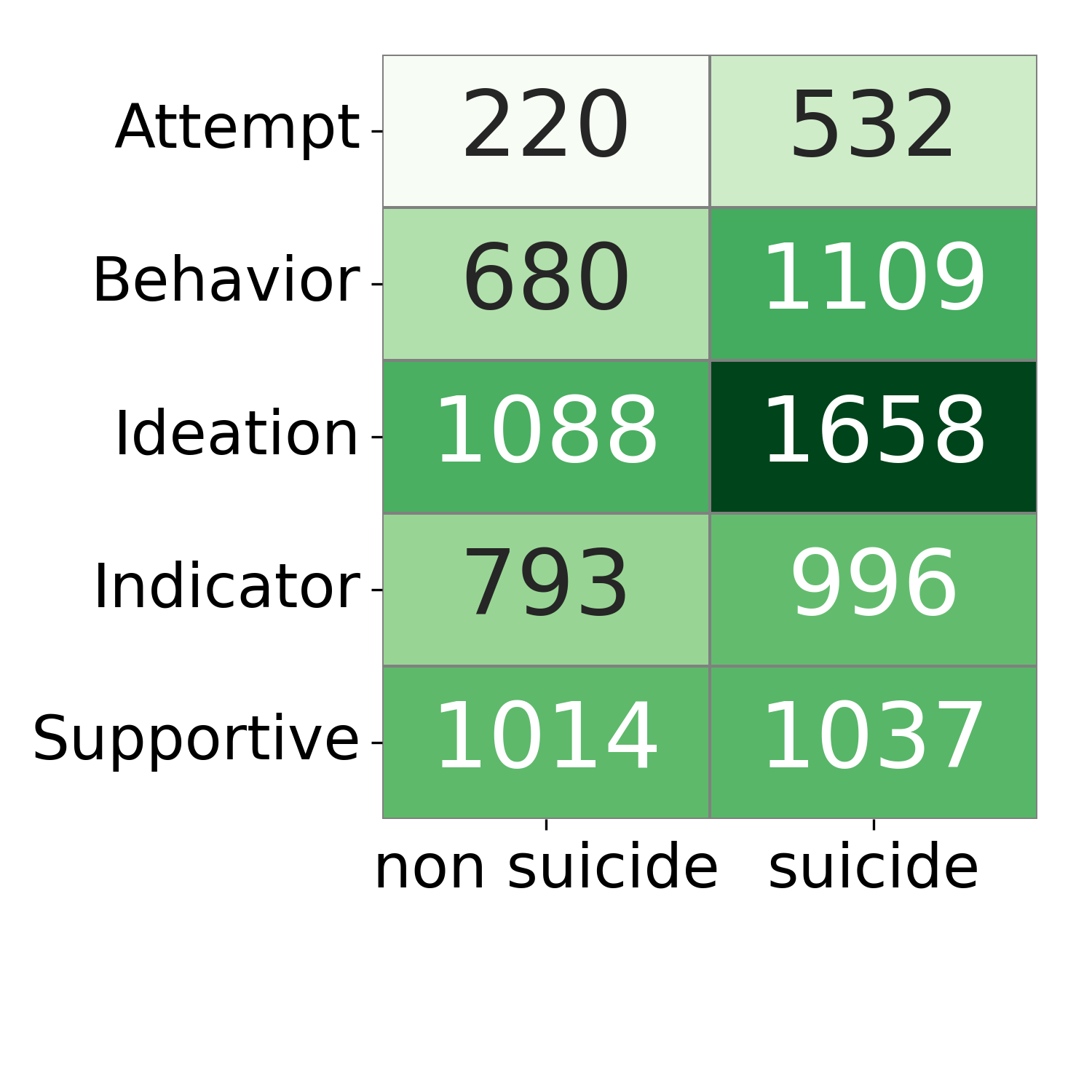}
    \caption{SENTINET}
    \label{fig:sentinet-cssrs}
    \end{subfigure}
    \hspace{-1em}
    \begin{subfigure}[b]{0.26\linewidth}
        \includegraphics[height=.85in, width=\linewidth]{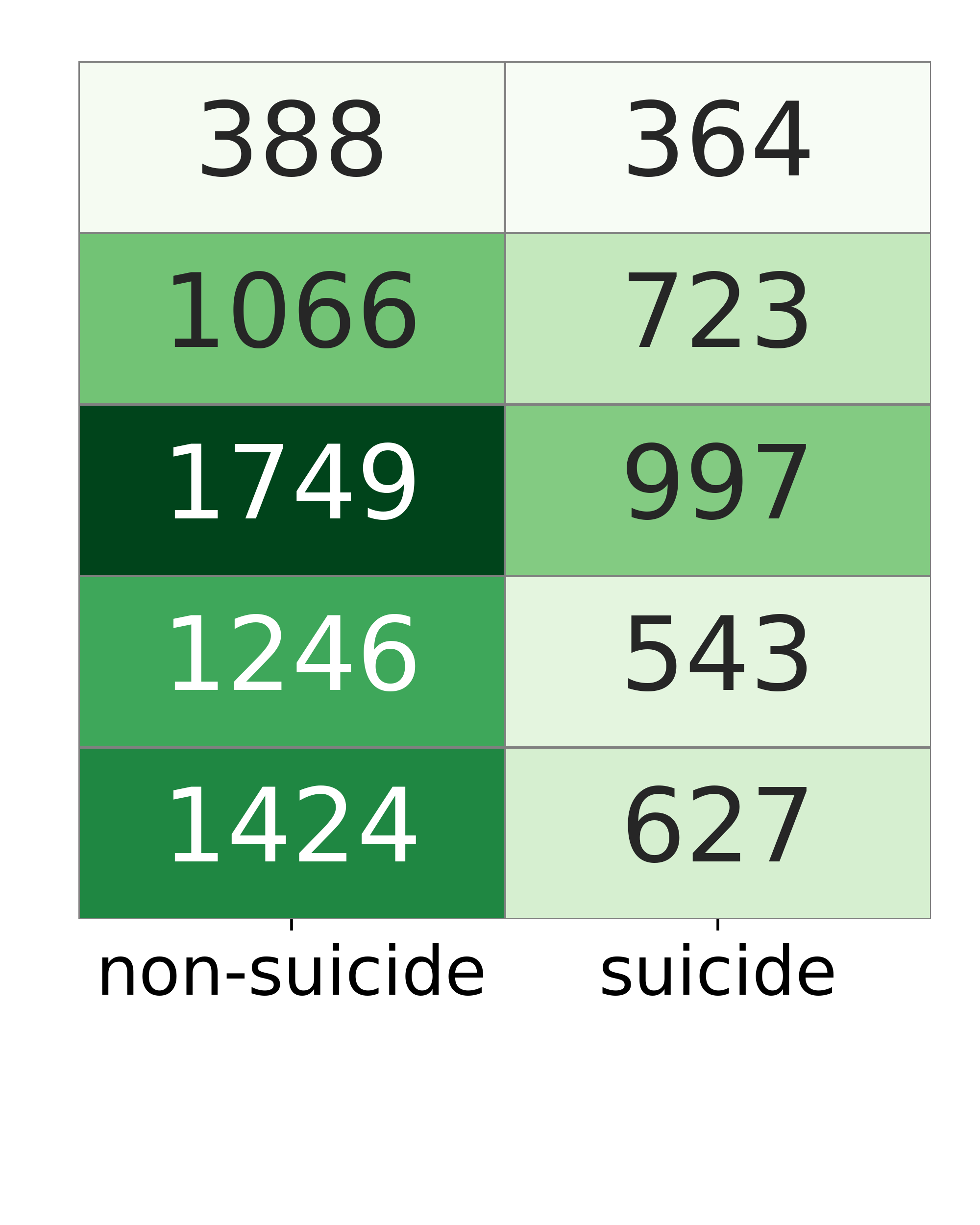}
        \caption{RoBERTa}
        \label{fig:roberta-cssrs}
    \end{subfigure}
    \begin{subfigure}[b]{\linewidth}
        \centering
        \includegraphics[width=\linewidth]{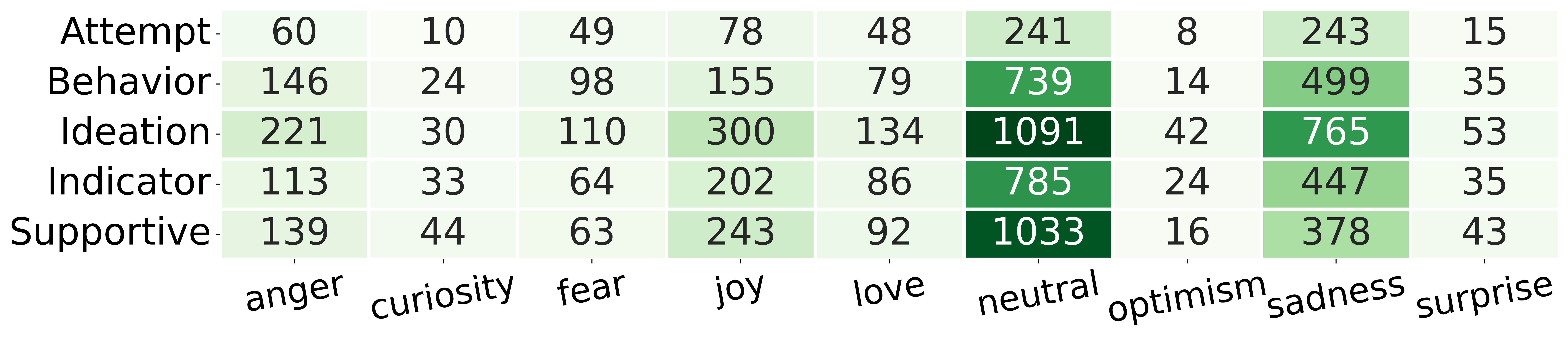}
        \caption{BERT-Emo}
        \label{fig:BERT-Emo-cssrs}
    \end{subfigure}
    \begin{subfigure}[b]{\linewidth}
        \centering
        \includegraphics[width=\linewidth]{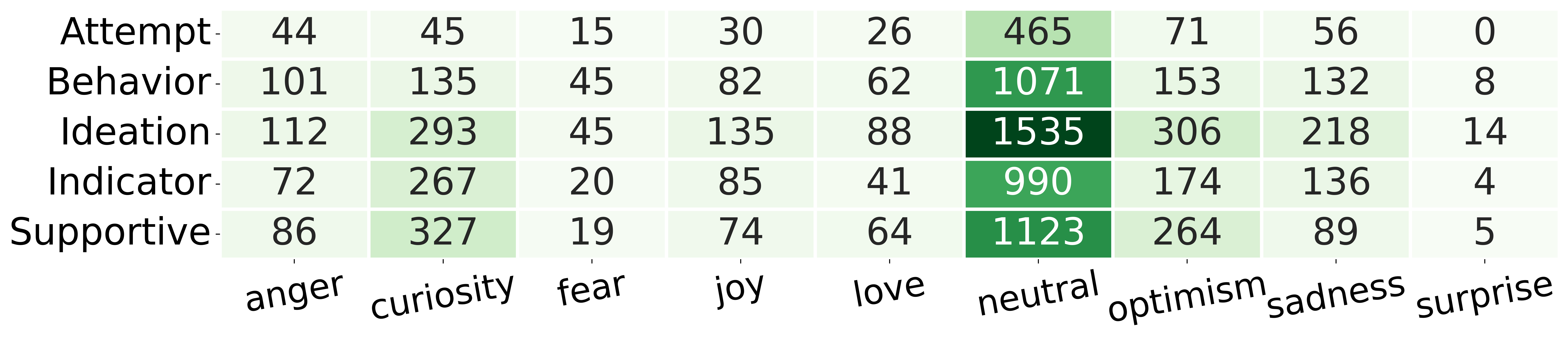}
        \caption{EmoRoBERTa}
        \label{fig:EmoRoBERTa-cssrs}
    \end{subfigure}
    \caption{Confusion matrices of C-SSRS dataset labels and fine-tuned transformer predictions.\vspace{-15pt}}
    \label{fig:cssrs_res}
\end{figure}

Assuming that the C-SSRS dataset annotated by four experts has gold-standard labels, implementing the four transformers on this collection becomes an evaluation of the models. The first confusion matrix in Fig.~\ref{fig:sentinet-cssrs} shows that SENTINET correctly detected suicidal ideation in the majority of the Attempt and Behavior posts. Meanwhile, its predictions disagree with the Ideation, Supportive, and Indicator labels. After merging the two labels that lack suicidal ideation, Supportive and Indicator, as non-suicidal, and the remaining labels as the suicidal class, we computed macro F1 and AUC scores of the SENTINET model in Table~\ref{tab:suicide_res}. However, the scores are only marginally better than random classification performance. This result is further significant considering how the C-SSRS dataset was a part of the SENTINET model's dataset. The second suicidal ideation detection transformer, RoBERTa, on the other hand, classified the majority of the posts of all labels as non-suicidal in Fig.~\ref{fig:roberta-cssrs}. The scores in Table~\ref{tab:suicide_res} show worse than random classification in terms of F1 score, while despite the wide range of classification thresholds AUC tries, results are also random. Thus, based on the assumption of the gold-standard dataset, we can claim that we cannot trust SENTINET or the more popular (with a higher download rate) RoBERTa models in suicidal ideation detection.

Subsequently, we detect emotions using BERT-Emo and EmoRoBERTa. The former transformer's confusion matrix in Fig.~\ref{fig:BERT-Emo-cssrs} mostly detects neutral emotions from all posts independent of their labels, together with sadness, joy, and anger. The next confusion matrix in Fig.~\ref{fig:EmoRoBERTa-cssrs} also mostly detects neutral emotions, with high curiosity and optimism in Ideation, Indicator, and Supportive posts. Since we selected to demonstrate only the overlapping emotions between the two models, we could compare their predictions with a 0.215 MCC score, which is closer to 0 than +1, indicating high disagreement.

\begin{figure}[t!]
    \centering
    \begin{subfigure}[b]{0.35\linewidth}
        \centering
        \includegraphics[height=.7in, width=\linewidth]{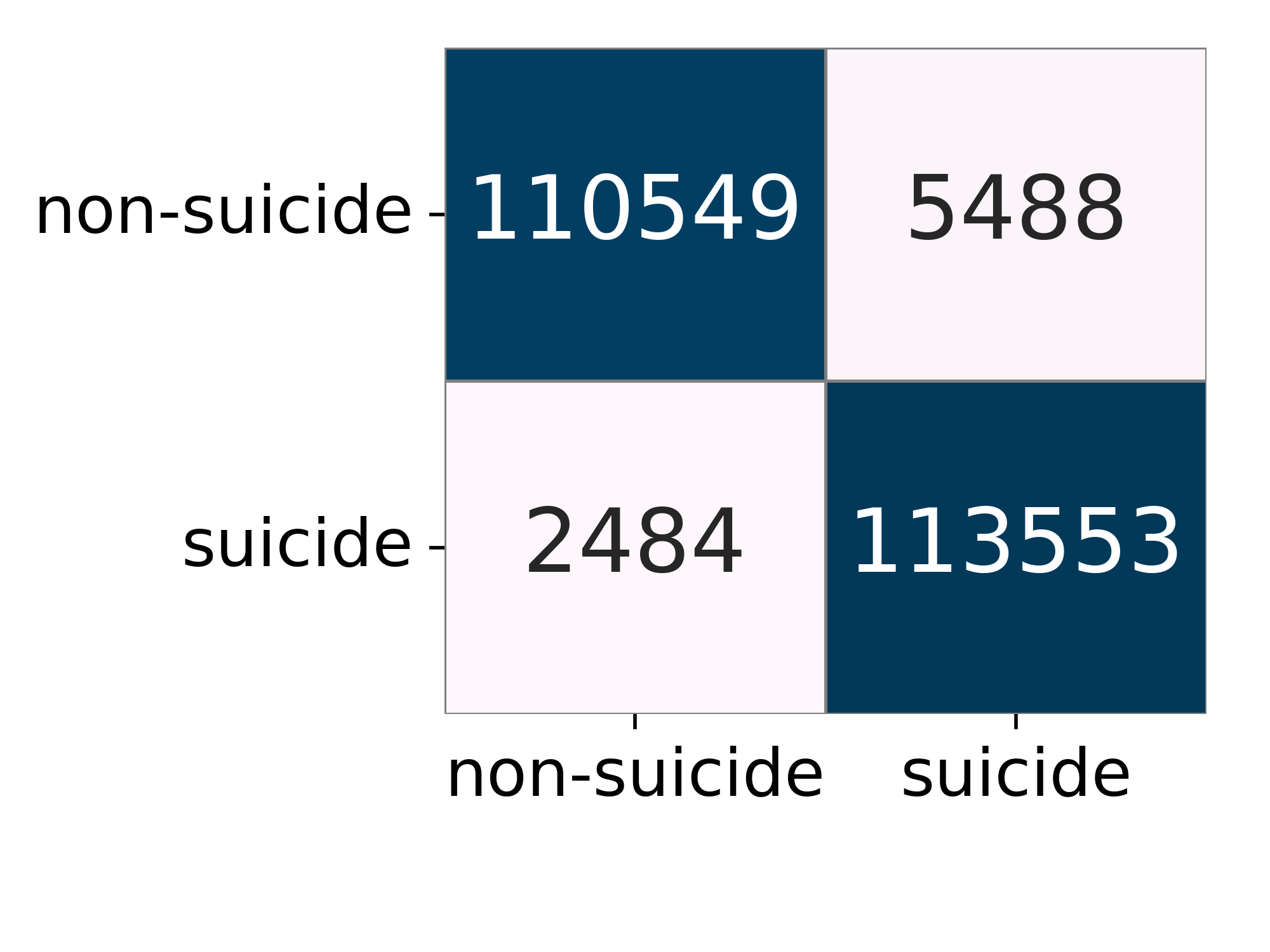}
        \caption{SENTINET}
        \label{fig:sentinet-sdd}
    \end{subfigure}
    \begin{subfigure}[b]{0.29\linewidth}
        \centering
        \includegraphics[height=.7in, width=\linewidth]{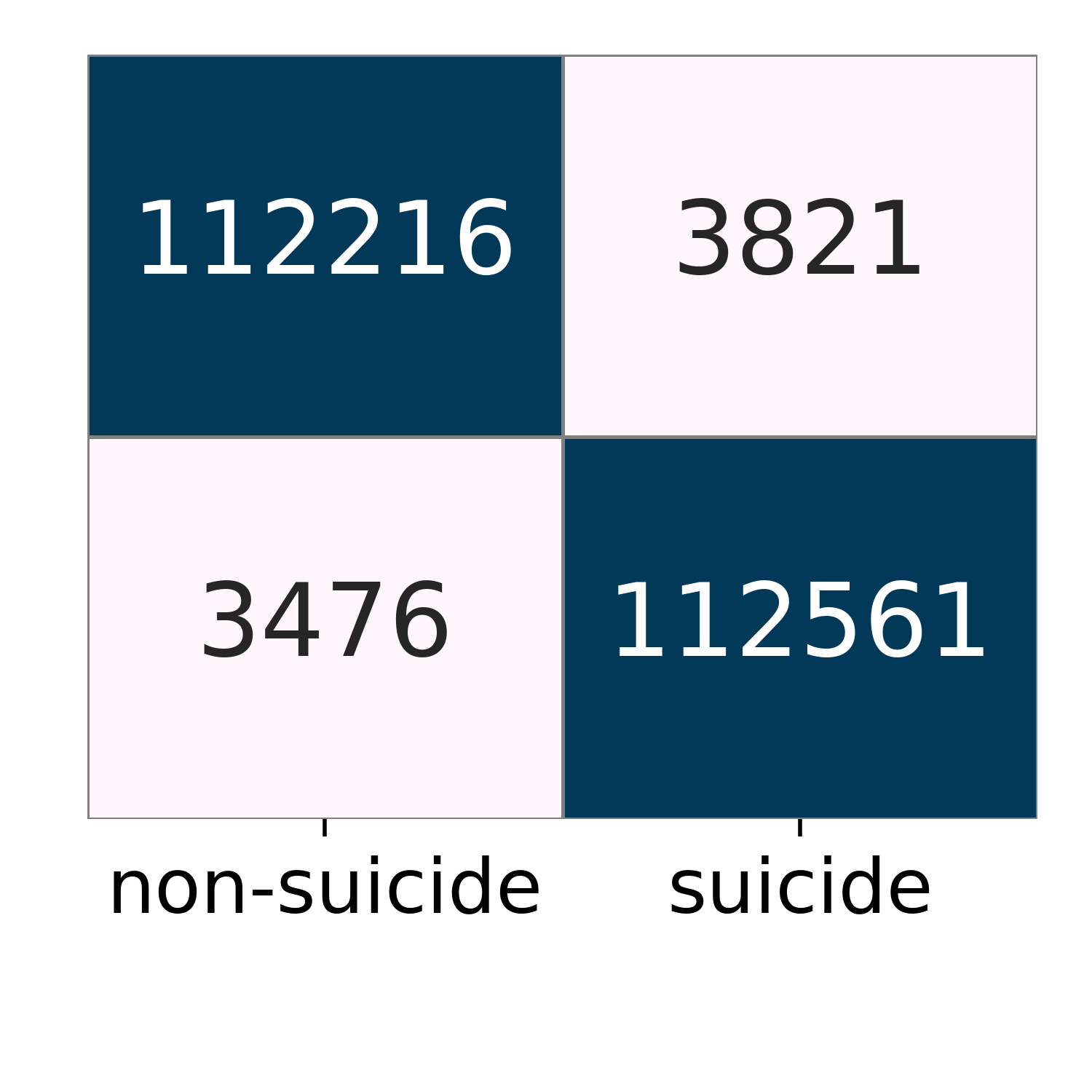}
        \caption{RoBERTa}
        \label{fig:roberta-sdd}
    \end{subfigure}
    \begin{subfigure}[b]{\linewidth}
        \centering
        \includegraphics[width=\linewidth]{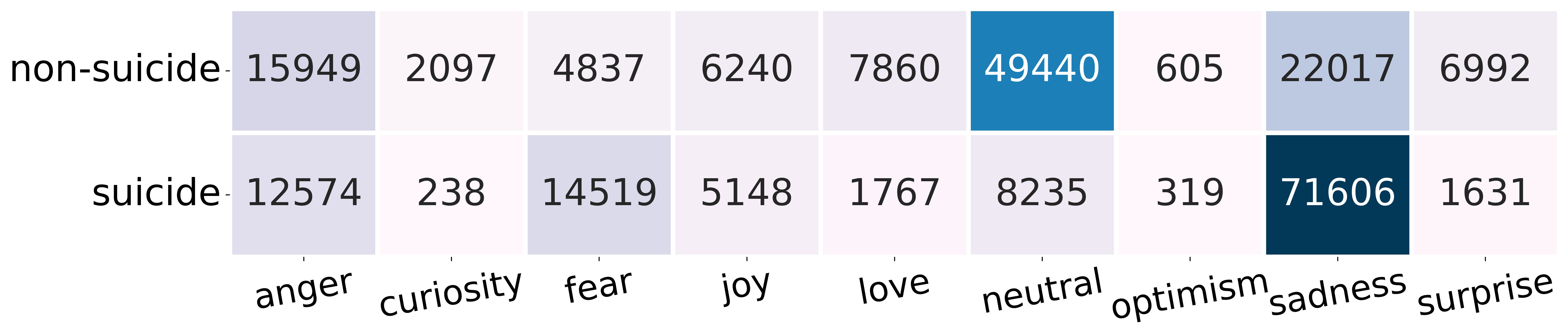}
        \caption{BERT-Emo}
        \label{fig:BERT-Emo-sdd}
    \end{subfigure}
    \begin{subfigure}[b]{\linewidth}
        \centering
        \includegraphics[width=\linewidth]{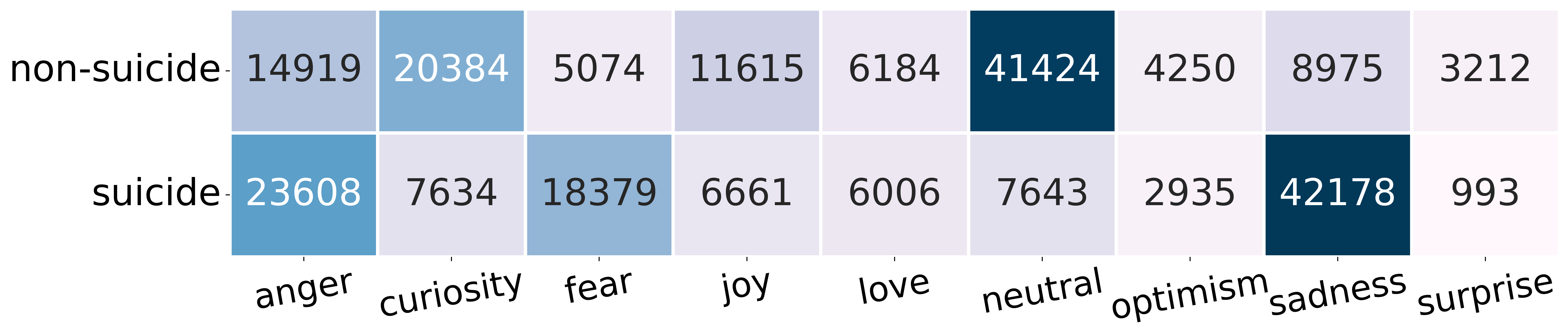}
        \caption{EmoRoBERTa}
        \label{fig:EmoRoBERTa-sdd}
    \end{subfigure}
    \caption{Confusion matrices of SDD dataset labels and fine-tuned transformer predictions.}
    \label{fig:SDD_res}
\end{figure}

The second English dataset posts, SDD, were labeled based on which subreddit they were from. The confusion matrices of the suicidal ideation detection transformers on this dataset in Fig.~\ref{fig:sentinet-sdd} and Fig.~\ref{fig:roberta-sdd} both show near-perfect results. When we enumerate their performance, Table~\ref{tab:suicide_res} shows 96\% F1 and 99\% AUC scores from both models. Considering the narrow overlap between the SDD dataset and these transformers, these high scores are concerning. Meantime, the two emotion detection models return similar results, where suicidal posts mainly have sadness, anger, and fear, and non-suicidal posts have neutral emotions and anger, except for high curiosity detected by EmoRoBERTa. Comparing the two emotion models based on the overlapping emotions returned a 0.380 MCC score, indicating moderate agreement.

\begin{figure}[t!]
    \centering
    \begin{subfigure}[b]{0.37\linewidth}
        \centering
        \includegraphics[height=.82in, width=\linewidth]{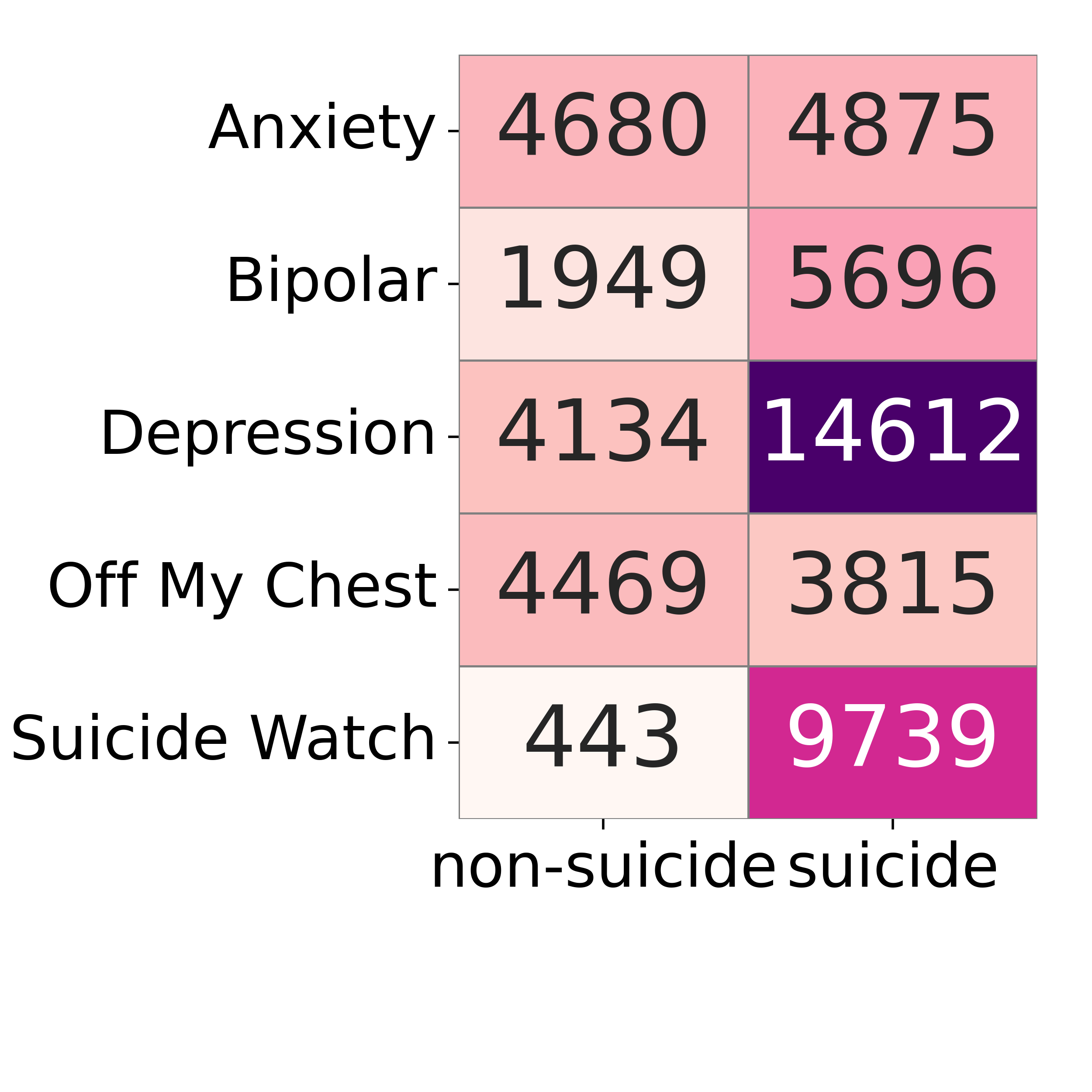}
        \caption{SENTINET}
        \label{fig:sentinet-swmh}
    \end{subfigure}
    \begin{subfigure}[b]{0.25\linewidth}
        \centering
        \includegraphics[height=.82in, width=\linewidth]{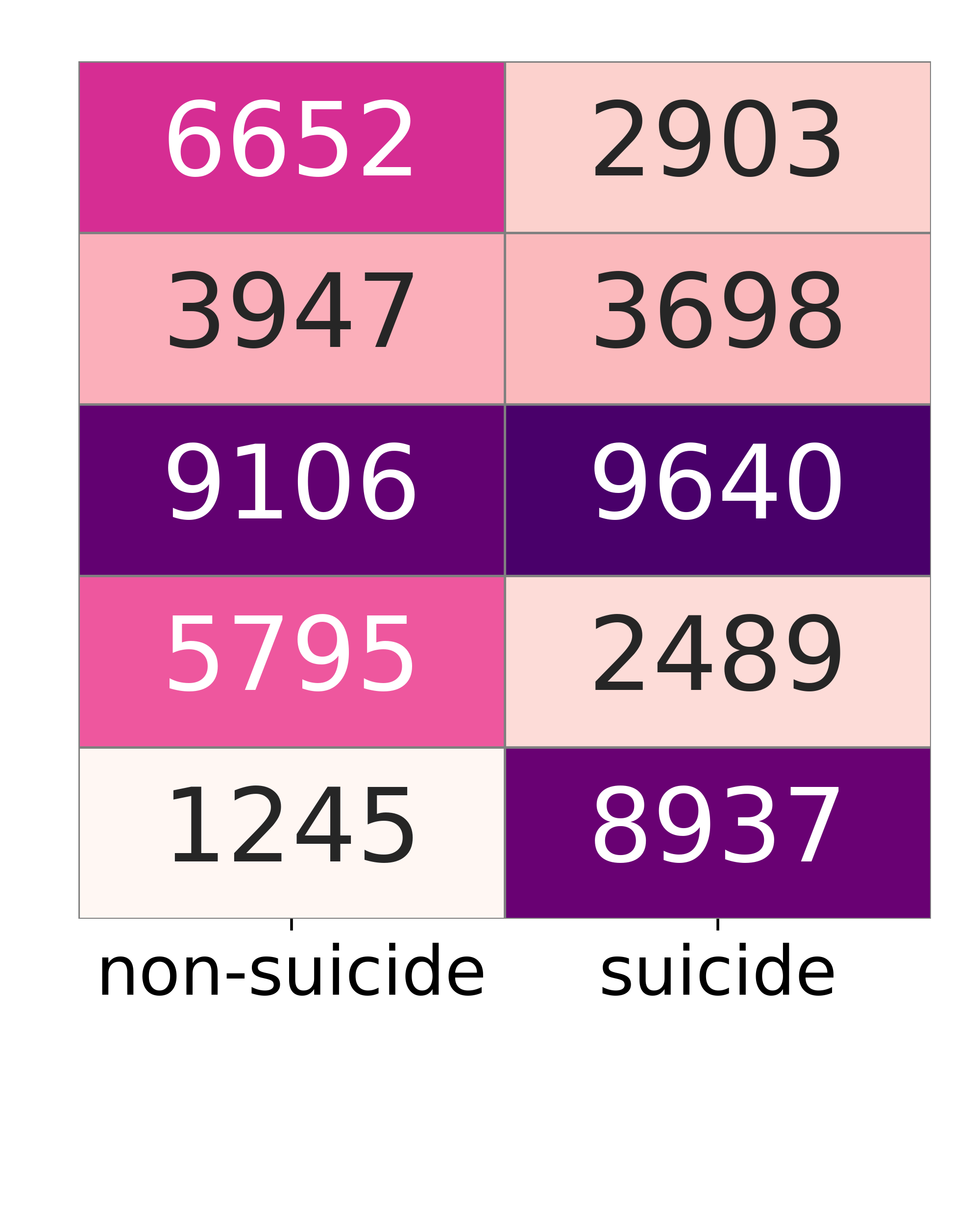}
        \caption{RoBERTa}
        \label{fig:roberta-swmh}
    \end{subfigure}
    \begin{subfigure}[b]{\linewidth}
        \centering
        \includegraphics[width=\linewidth]{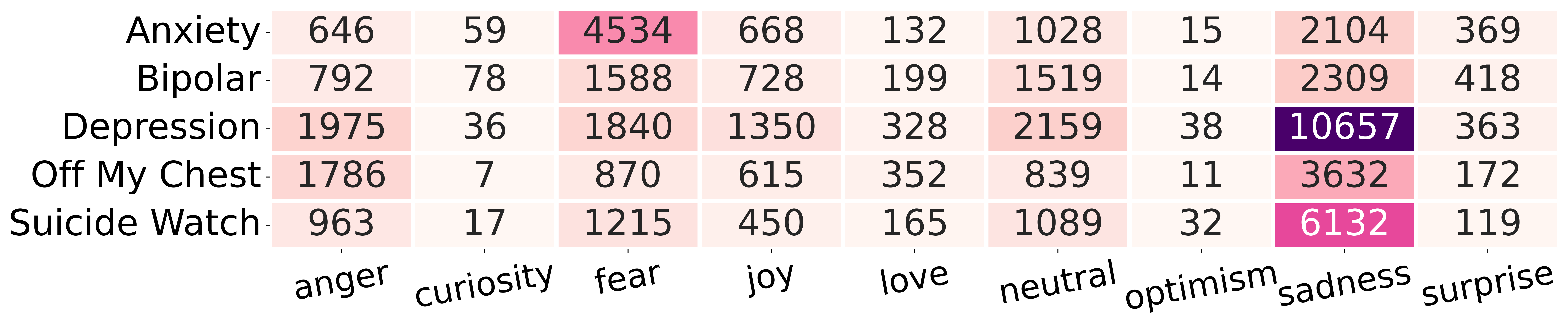}
        \caption{BERT-Emo}
        \label{fig:BERT-Emo-SWMH}
    \end{subfigure}
    \begin{subfigure}[b]{\linewidth}
        \centering
        \includegraphics[width=\linewidth]{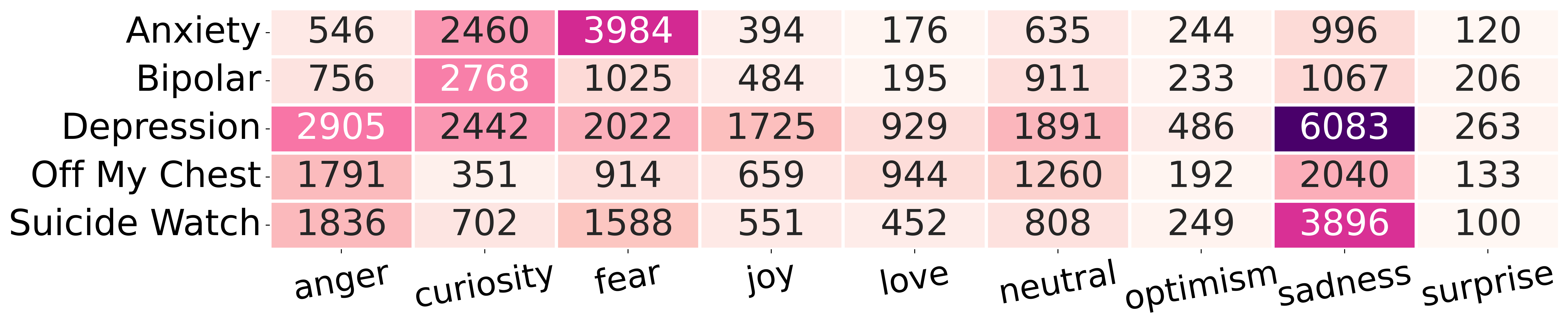}
        \caption{EmoRoBERTa}
        \label{fig:EmoRoBERTa-SWMH}
    \end{subfigure}
    \caption{Confusion matrices of SWMH dataset labels and fine-tuned transformer predictions.\vspace{-16pt}}
    \label{fig:SWMH_res}
\end{figure}

Finally, the SWMH dataset, which was also labeled by the posts' subreddits, returned two different confusion matrices from the suicidal ideation detection models as seen in Fig.~\ref{fig:sentinet-swmh} and Fig.~\ref{fig:roberta-swmh}. While both models correctly detect posts from SuicideWatch, SENTINET classifies Depression posts as suicidal, while RoBERTa classifies only half of them as suicidal. Considering the scenario where we consider SuicideWatch as the suicidal class and the remaining posts as non-suicidal, we compute the performance of the two transformers in Table~\ref{tab:suicide_res} with random F1 but above 80\% AUC scores. Further, the two emotion detection models in Fig.~\ref{fig:SWMH_res} return sadness from Depression and SuicideWatch, with a 0.323 MCC score.

\section{Discussion}

\subsection{Reliable Annotation}
In the present study, we addressed several gaps in suicide-related NLP research by constructing a novel high-quality Turkish suicidal ideation dataset. We introduced a resource-conscious, multi-level annotation framework. Two researchers annotated the entire dataset; two large language models (LLMs) served as tiebreakers for non-sensitive cases, where a sensitive case involved disagreement about the presence of suicidal ideation in a post. In these cases, we relied exclusively on a third expert annotator to ensure reliability while preserving resource efficiency. Additionally, we introduced a controlled, non-harmful degree of label noise via the LLMs to encourage model generalization and prevent overfitting in future training. Inter-annotator agreement metrics confirmed the robustness of our annotation process.

Due to ethical and legal constraints, we do not release the dataset. However, we provide detailed information throughout the present paper, including the annotation methodology, to ensure transparency and reproducibility. We offer this pipeline as a blueprint for building high-reliability mental health datasets in all languages.

\subsection{Model Failures - Absent Annotations}
Despite the quality of our Turkish dataset and the extensive annotation design, the selected four Turkish transformer models failed to capture different sentiments and emotions between suicidal and non-suicidal posts. This failure cannot be attributed to poor annotation quality as our dataset is gold-standard. Instead, it highlights the limitations of Turkish transformer models: The fine-tuned and available models require more fine-tuning on labeled Turkish data collections for accurate detection of nuanced, context-sensitive mental health signals.

\subsection{Toward Better Practices}
Moreover, the challenge of needing more reliably labeled data is not unique to Turkish. Our evaluation of three English suicidal ideation datasets revealed the same trend. Despite being included in the fine-tuning set of the SENTINET model, the C-SSRS dataset (with gold-standard annotations) was classified with near-random performance by two suicidal ideation detection transformers. In contrast, the same transformers performed significantly better on auto-labeled SDD and SWMH despite having minimal direct overlap with their training sets. This difference suggests that the models are learning superficial patterns such as subreddit identifiers (e.g., titles, which were the labeling factor) or community-specific phrasing rather than the underlying cues of suicidal ideation. This observation calls for questioning the two common approaches: (1) using auto-labeled datasets for suicidal ideation detection and (2) using off-the-shelf, fine-tuned models for mental health detection without validating their training and fine-tuning sets. Studies that use auto-labeled datasets or the models developed on them need to avoid claiming to perform any form of suicidal ideation detection. What they provide is source attribution: Identifying the subreddit origin of a post based on its textual content. We do not condemn auto-labeled datasets, as they are essential for pre-training transformer models for introducing social media context. However, the field must also embrace (1) questioning and revising existing benchmarks and (2) creating more gold-standard datasets using scalable annotation frameworks such as ours.

\subsection{Ethical Implications}
We designed this study with ethical integrity as a central concern, especially given the sensitivity of suicidal ideation as a research domain. Our data collection was limited to publicly available social media content. Our annotation protocol explicitly avoided having non-experts make final decisions on potentially harmful content. Additionally, the controlled introduction of label noise through LLMs was carefully designed to avoid ethical concerns, so no false suicidal labels were assigned, and this step was restricted to non-sensitive instances.

We emphasize that while the suicidal ideation detection systems we develop are important for the future of suicide prevention, they cannot and will not replace clinical judgment. Rather, we aim to encourage more individuals to receive help from mental health professionals. Finally, we advocate for transparency in model training pipelines and dataset construction practices in mental health NLP. Complete reliance on scale, speed, or convenience should never come at the cost of data or model reliability.

\section*{Acknowledgements}
This work was supported by the Scientific and Technological Research Council of Türkiye (TÜBİTAK) under Grant \#124E132.

\section*{References}

\bibliographystyle{ieeetr}
\bibliography{references}

\begin{thebibliography}{10}

\bibitem{who_suicide_2025}
{World Health Organization}, ``Suicide,'' Mar. 2025.
\newblock Accessed: 2025-05-15.

\bibitem{kessler2005prevalence}
R.~C. Kessler, O.~Demler, R.~G. Frank, M.~Olfson, H.~A. Pincus, E.~E. Walters, P.~Wang, K.~B. Wells, and A.~M. Zaslavsky, ``Prevalence and treatment of mental disorders, 1990 to 2003,'' {\em New England Journal of Medicine}, vol.~352, no.~24, pp.~2515--2523, 2005.

\bibitem{kazdin2011evidence}
A.~E. Kazdin, ``Evidence-based treatment research: Advances, limitations, and next steps.,'' {\em American Psychologist}, vol.~66, no.~8, p.~685, 2011.

\bibitem{ji2020suicidal}
S.~Ji, S.~Pan, X.~Li, E.~Cambria, G.~Long, and Z.~Huang, ``Suicidal ideation detection: A review of machine learning methods and applications,'' {\em IEEE Transactions on Computational Social Systems}, vol.~8, no.~1, pp.~214--226, 2020.

\bibitem{riera2024clinical}
P.~Riera-Serra, G.~Navarra-Ventura, A.~Castro, M.~Gili, A.~Salazar-Cedillo, I.~Ricci-Cabello, L.~Rold{\'a}n-Esp{\'\i}nola, V.~Coronado-Simsic, M.~Garc{\'\i}a-Toro, R.~G{\'o}mez-Juanes, {\em et~al.}, ``Clinical predictors of suicidal ideation, suicide attempts and suicide death in depressive disorder: a systematic review and meta-analysis,'' {\em European archives of psychiatry and clinical neuroscience}, vol.~274, no.~7, pp.~1543--1563, 2024.

\bibitem{pigoni2024machine}
A.~Pigoni, G.~Delvecchio, N.~Turtulici, D.~Madonna, P.~Pietrini, L.~Cecchetti, and P.~Brambilla, ``Machine learning and the prediction of suicide in psychiatric populations: a systematic review,'' {\em Translational psychiatry}, vol.~14, no.~1, p.~140, 2024.

\bibitem{coppersmith2018natural}
G.~Coppersmith, R.~Leary, P.~Crutchley, and A.~Fine, ``Natural language processing of social media as screening for suicide risk,'' {\em Biomedical informatics insights}, vol.~10, p.~1178222618792860, 2018.

\bibitem{shukla2025enhancing}
S.~S.~P. Shukla and M.~P. Singh, ``Enhancing suicidal ideation detection through advanced feature selection and stacked deep learning models,'' {\em Applied Intelligence}, vol.~55, no.~4, p.~303, 2025.

\bibitem{gorai2024bert}
J.~Gorai and D.~K. Shaw, ``A bert-encoded ensembled cnn model for suicide risk identification in social media posts,'' {\em Neural Computing and Applications}, vol.~36, no.~18, pp.~10955--10970, 2024.

\bibitem{qorich2024advanced}
M.~Qorich and R.~El~Ouazzani, ``Advanced deep learning and large language models for suicide ideation detection on social media,'' {\em Progress in Artificial Intelligence}, vol.~13, no.~2, pp.~135--147, 2024.

\bibitem{yakar2017suicide}
M.~Yakar, K.~Temur{\c{c}}in, and I.~Kervankiran, ``Suicide in {T}urkey: its changes and regional differences,'' {\em Bulletin of Geography. Socio-economic Series}, no.~35, pp.~123--143, 2017.

\bibitem{erenler2023suicide}
A.~K. Erenler, B.~G{\"u}m{\"u}{\c{s}}, and M.~Y{\i}lmaz, ``Suicide deaths in {T}urkey: A nationwide study between 2008 and 2018,'' {\em Gaziosmanpa{\c{s}}a {\"U}niversitesi T{\i}p Fak{\"u}ltesi Dergisi}, vol.~14, no.~3, pp.~42--155, 2023.

\bibitem{yildiz2023suicide}
M.~Y{\i}ld{\i}z, K.~D. Batun, H.~{\c{S}}ahino{\u{g}}lu, M.~S. Ery{\i}lmaz, B.~{\"O}zel, B.~Atao{\u{g}}lu, and S.~E. H{\i}d{\i}ro{\u{g}}lu, ``Suicide among doctors in {T}urkey: Differences across gender, medical specialty and the method of suicide,'' {\em Advances in Clinical and Experimental Medicine}, vol.~32, no.~9, pp.~977--986, 2023.

\bibitem{yucel2023suicidal}
Y.~Y{\"u}cel and B.~Kabalay, ``Suicidal economy of {T}urkey in times of crisis: 2018 crisis and beyond,'' {\em Critical Sociology}, vol.~49, no.~4-5, pp.~783--800, 2023.

\bibitem{gaur2019knowledge}
M.~Gaur, A.~Alambo, J.~P. Sain, U.~Kursuncu, K.~Thirunarayan, R.~Kavuluru, A.~Sheth, R.~Welton, and J.~Pathak, ``Knowledge-aware assessment of severity of suicide risk for early intervention,'' in {\em The world wide web conference}, pp.~514--525, 2019.

\bibitem{desu2022suicide}
V.~Desu, N.~Komati, S.~Lingamaneni, and F.~Shaik, ``Suicide and depression detection in social media forums,'' in {\em Smart Intelligent Computing and Applications, Volume 2: Proceedings of Fifth International Conference on Smart Computing and Informatics (SCI 2021)}, pp.~263--270, Springer, 2022.

\bibitem{ji2021suicidal}
S.~Ji, X.~Li, Z.~Huang, and E.~Cambria, ``Suicidal ideation and mental disorder detection with attentive relation networks,'' {\em Neural Computing and Applications}, 2021.

\bibitem{ezerceli2024mental}
{\"O}.~Ezerceli and R.~Dehkharghani, ``Mental disorder and suicidal ideation detection from social media using deep neural networks,'' {\em Journal of Computational Social Science}, vol.~7, no.~3, pp.~2277--2307, 2024.

\bibitem{turk2023predicting}
B.~Turk and H.~H. Tali, ``Predicting suicidal thoughts in a non-clinical sample using machine learning methods,'' {\em Dusunen Adam}, vol.~36, no.~3, pp.~179--188, 2023.

\bibitem{alshammari2024mental}
Q.~Alshammari and S.~Aky{\"u}z, ``Mental health on twitter in turkey: Sentiment analysis with transformers,'' in {\em Decision Making in Healthcare Systems}, pp.~391--402, Springer, 2024.

\bibitem{sari2025vaccine}
S.~Sari and U.~Bayram, ``Vaccine hesitancy in {T}{\"u}rkiye: A natural language processing study on social media,'' {\em Turkish Journal of Electrical Engineering and Computer Sciences}, vol.~33, no.~3, pp.~392--409, 2025.

\bibitem{richardson2007beautiful}
L.~Richardson, ``Beautiful soup documentation,'' 2007.

\bibitem{aldeen2023chatgpt}
M.~Aldeen, J.~Luo, A.~Lian, V.~Zheng, A.~Hong, P.~Yetukuri, and L.~Cheng, ``Chatgpt vs. human annotators: A comprehensive analysis of chatgpt for text annotation,'' in {\em 2023 International Conference on Machine Learning and Applications (ICMLA)}, pp.~602--609, IEEE, 2023.

\bibitem{ezin2024using}
E.~Ezin, R.~S. Kiziltepe, and M.~Karakus, ``Using llms for annotation and ml methods for comparative analysis of large-scale turkish sentiment datasets,'' in {\em 2024 9th International Conference on Computer Science and Engineering (UBMK)}, pp.~204--209, IEEE, 2024.

\bibitem{nasution2024chatgpt}
A.~H. Nasution and A.~Onan, ``Chatgpt label: Comparing the quality of human-generated and llm-generated annotations in low-resource language nlp tasks,'' {\em IEEE Access}, 2024.

\bibitem{naseem2022hybrid}
U.~Naseem, M.~Khushi, J.~Kim, and A.~G. Dunn, ``Hybrid text representation for explainable suicide risk identification on social media,'' {\em IEEE transactions on computational social systems}, 2022.

\bibitem{soun2024rise}
R.~S. Soun, A.~T. Neerkaje, R.~Sawhney, N.~Aletras, and P.~Nakov, ``Rise: Robust early-exiting internal classifiers for suicide risk evaluation,'' in {\em Proceedings of the 2024 Joint International Conference on Computational Linguistics, Language Resources and Evaluation (LREC-COLING 2024)}, pp.~14134--14145, 2024.

\bibitem{nikhileswar2021suicide}
K.~Nikhileswar, D.~Vishal, L.~Sphoorthi, and S.~Fathimabi, ``Suicide ideation detection in social media forums,'' in {\em 2021 2nd International Conference on Smart Electronics and Communication (ICOSEC)}, pp.~1741--1747, IEEE, 2021.

\bibitem{lin2024data}
E.~Lin, J.~Sun, H.~Chen, and M.~H. Mahoor, ``Data quality matters: Suicide intention detection on social media posts using roberta-cnn,'' in {\em 2024 46th Annual International Conference of the IEEE Engineering in Medicine and Biology Society (EMBC)}, pp.~1--5, IEEE, 2024.

\bibitem{stoica2024pearson}
P.~Stoica and P.~Babu, ``Pearson--matthews correlation coefficients for binary and multinary classification,'' {\em Signal Processing}, vol.~222, p.~109511, 2024.

\end{thebibliography}

\end{document}